\documentclass[5p, times]{elsarticle}

\usepackage{amssymb}
\usepackage{bbm}
\usepackage{setspace}
\pdfoutput=1

\usepackage{times}
\usepackage{soul}
\usepackage{url}
\usepackage[hidelinks]{hyperref}
\usepackage[utf8]{inputenc}
\usepackage[small]{caption}
\usepackage{graphicx}
\usepackage{amsmath}
\usepackage{mathrsfs}
\usepackage{amsthm}
\usepackage{booktabs}
\usepackage{algorithm}
\usepackage{algorithmic}
\usepackage{pgfplots}
\usepackage{tikz}
\usepackage{sansmath}
\usetikzlibrary{shadings,intersections}
\usetikzlibrary{calc}
\usepackage{latexsym}
\usepackage{amsfonts}
\usepackage{colortbl}
\usepackage{tikz}
\usetikzlibrary{arrows,shapes,chains}
\usepackage{subfig}

\newtheorem{prop}{Proposition}

\journal{Journal of \LaTeX\ Templates}

\begin{document}

\begin{frontmatter}

\title{M\"{o}biusE: Knowledge Graph Embedding on M\"{o}bius Ring}
%\tnotetext[mytitlenote]{Fully documented templates are available in the elsarticle package on \href{http://www.ctan.org/tex-archive/macros/latex/contrib/elsarticle}{CTAN}.}

%% Group authors per affiliation:
%\author{Elsevier\fnref{myfootnote}}
%\address{Radarweg 29, Amsterdam}
%\fntext[myfootnote]{Since 1880.}

%% or include affiliations in footnotes:
\author[add1]{Yao Chen\corref{mycorrespondingauthor}}
\cortext[mycorrespondingauthor]{Corresponding author}
\ead{chenyao@swufe.edu.cn}
%\ead[url]{www.elsevier.com}
\author[add1]{Jiangang Liu}
\author[add1]{Zhe Zhang}
\author[add2]{Shiping Wen}
\author[add1]{Wenjun Xiong}

\address[add1]{Department of Computer Science, Southwestern University of Finance and Economics, China}
\address[add2]{Centre for Artificial Intelligence, University of Technology Sydney, Sydney, Australia}

\begin{abstract}
In this work, we propose a novel Knowledge Graph Embedding (KGE) strategy, called M\"{o}biusE, in which the entities and relations are embedded to the surface of a M\"{o}bius ring. The proposition of such a strategy is inspired by the classic TorusE, in which the addition of two arbitrary elements is subject to a modulus operation. In this sense, TorusE naturally guarantees the critical boundedness of embedding vectors in KGE. However, the nonlinear property of addition operation on Torus ring is uniquely derived by the modulus operation, which in some extent restricts the expressiveness of TorusE. As a further generalization of TorusE, M\"{o}biusE also uses modulus operation to preserve the closeness of addition operation on it, but the coordinates on M\"{o}bius ring interacts with each other in the following way: {\em \color{red} any vector on the surface of a M\"{o}bius ring moves along its parametric trace will goes to the right opposite direction after a cycle}. Hence, M\"{o}biusE assumes much more nonlinear representativeness than that of TorusE, and in turn it generates much more precise embedding results. In our experiments, M\"{o}biusE outperforms TorusE and other classic embedding strategies in several key indicators.  
\end{abstract}

\begin{keyword}
M\"{o}bius ring\sep Torus ring \sep Knowledge graph \sep Embedding
%\MSC[2010] 00-01\sep  99-00
\end{keyword}

\end{frontmatter}

% \linenumbers

%---------------------------------------------------------------------------------------------------------%
\section{Introduction}
Graph or network structure can usually be used to model the intrinsic information behind data \cite{lu2018embedding}. As an typical example, a knowledge graph (KG) is a collection of facts of the real world, related examples include DBpedia \cite{auer07dbpedia}, YAGO \cite{suchanek07yago} and Freebase \cite{kurt08freebase}. These built KG database can be applied in many realistic engineering tasks, such as knowledge inference, question answering, and sample labeling. Driven by realistic applications, an intriguing and challenging problem for KG is: how to infer those unknown knowledge (or missing data) via the existing built KG database. 

For any KG, the basic element of knowledge is represented by a triplet $(h, r, t)$, which composed of two entities $h, t$ and one relation $r$. In detail, $h$ is called the head, $r$ is called the tail, and $(h, r, t)$ means head $h$ maps to tail $t$ via the operation of relation $r$. For example, the entity $h=$ {\tt U.S.A} projects to $t=$ {\tt DonaldTrump} under the relation $h=$ {\tt ThePresidentOf}, and the entity $h=$ {\tt U.S.A} projects to $t=$ {\tt MikePence} under the relation $h=$ {\tt TheVicePresidentOf}. Based on these two triplets, and considering the fact of ``the president and vice president of a country at the same time are colleagues'', one predicts the existence of the triplet $(${\tt DonaldTrump}, {\tt IsColleagueOf}, {\tt MikePence}$)$ if such triplet is not contained in the given database. For realistic applications such as question answering, the core task is to predict those triplets which are not listed in the given database, such as task is called link prediction in KG. 

There exists many types of method for link prediction in KG, among which translation-based methods treats each entity and relation as embedded vectors in some constructed algebraic space, and each true triplet can be translated to a simple constraint described by an algebraic expression $f(h,r,t)$. Using such a method, if some $(h, r, t)$ is not presented in KG but the corresponding algebraic constraint index $f(h,r,t)$ is very small, then it is reasonable to believe that $(h,r,t)$ is a missing fact. Due to the choosing of different algebraic space, many different translation-based models have be obtained. TransE \cite{bordes2013translating} is the first translation-based model for link prediction tasks, in which it uses the Euclidean space $\mathbb{R}^n$ as the embedding space and the constraint quantity for each triplet is set as simple as $\|h+r-t\|$. TransR \cite{lin2015learning} builds the constraint equation by mapping the entities to a subspace derived by the relation vector. TansH \cite{wang2014knowledge} improves TransE in dealing with reflexive/ONE-TO-MANY/MANY-TO-ONE/MANY-TO-MANY cases in KG, such an improvement is caused by treating each relation as a translating operation on a hyperplane. DistMult \cite{yang2015embedding} utilizes a simple bilinear formulation to model the constraint generated by triplets, in this way the composition of relation is characterized by matrix multiplication, which makes such a model good at capturing relational semantics. ComplEx \cite{trouillon2012complex} uses the standard dot product between embeddings and demonstrates the capability of complex-valued embeddings instead of real-valued numbers. TorusE \cite{ebisu2018toruse} chooses a compact Lie group, a torus, as the embedding space, and is more scalable to large-size knowledge graphs because of its lower computational complexity. Other types of embedding methods include TransA \cite{jia2018knowledge}, a locally and temporally adaptive translation-based approach; ConvE \cite{dettmers2018convolutional}, a multi-layer convolutional network embedding model; TransF, an embedding strategy with flexibility in each triplet constraint; QuaternionE \cite{zhang2019quaternion}, an embedding method treats each relation as a quaternion in hypercomplex space. 

Among the above embedding methods, the basic operation behind TorusE is a simple modulus operation and this operation automatically regularizes the calculated results to the given bounded area. In this way, there is no need to make an extra regularization operation in TorusE. Inspired by this property of TorusE, we move beyond one-dimensional modulus operation on Lie group and propose M\"{o}biusE, which takes a M\"{o}bius ring (whose basic dimension is two) as the embedding space. This M\"{o}biusE has the following major benefits for embedding tasks: At first, M\"{o}biusE provides much more flexibility since each point on the surface of a M\"{o}bius ring has two different expressions; Second, the expressiveness of M\"{o}biusE is much greater than TorusE due to the complex distance function built in M\"{o}bius ring; Third, M\"{o}biusE also automatically truncates the training vectors to constraint space as that of TorusE; Finally, M\"{o}biusE subsumes TorusE, and inherits all the attractive properties of TorusE, such as its ability to model symmetry/antisymmetry, inversion, and composition. 

The rest of the paper is organized as follows: Section \ref{sec:mobiusring} revisits the idea of TorusE and proposes the basic idea of M\"{o}biusE, especially, we discuss about how to define the distance function on a M\"{o}bius ring. Section \ref{sec:experiments} provides the experiment results for M\"{o}biusE on datasets FB15K and WN18. Section \ref{sec:relatedworks} discuss about related works and compare M\"{o}biusE with other embedding methods. Section \ref{sec:conclusions} concludes this paper. In the end, Section \ref{sec:appendix} lists the proof for several key propositions on M\"{o}bius ring. 

\begin{small}
	\begin{table*}\label{tab:TypicalModels}
		\centering
		\begin{tabular}{ccc}
			\toprule
			Name & Triplet Score Function & Embedding Space\\
			\midrule
			TransE \cite{bordes2013translating} 	& $||\bf{h}+\bf{r}-\bf{t}||$ 	& $\bf{h,r,t}\in\mathbb{R}^n$\\
			TransH \cite{wang2014knowledge} 		& $||\bf{h} - \bf{w_r^\mathrm{T}hw_r}+d_r-(\bf{t} - \bf{w_r^\mathrm{T}tw_r})||$&
$\bf{h,t,d_r,w_r}\in\mathbb{R}^n$\\
			RESCAL \cite{nickel2011rescal} 			& $\bf{h}^T\bf{M}_r\bf{t}$ 		& $\|\bf{h}\|_2\leq 1, \|\bf{t}\|_2\leq 1, \|\bf{M}_r\|_F\leq 1$\\
			TorusE \cite{ebisu2018toruse} &$||\bf{h}+\bf{r}-\bf{t}||_{T^n}$&$\bf{h,r,t}\in \mathbb{T}^n$\\
			Distmult \cite{yang2015embedding} &$-\bf{h^\mathrm{T}}diag(\bf{r})\bf{t}$ &$\bf{h,r,t}\in\mathbb{R}^n$\\
			ComplEx \cite{trouillon2012complex} &$-{\rm Re}(\bf{h^\mathrm{T}diag(r)\bar{t}})$&$\bf{h,r,t}\in\mathbb{C}^n$\\
			M\"{o}biusE (ours) &${\rm dist}(\bf{h}\oplus\bf{r}, \bf{t})$&$\bf{h}, \bf{r}, \bf{t}\in\mathbb{M}^{q/p}_n$\\
			\bottomrule
		\end{tabular}
		\caption{Scoring Functions of Typical KGE Models}
		\label{models}
	\end{table*}
\end{small}

%RotatE \cite{sun2019rotate} &$||\bf{h}\circ \bf{r}-\bf{t}||$&$\bf{h,r,t}\in\mathbb{C}^n,~|r_i|=1$\\

%---------------------------------------------------------------------------------------------------------%
\section{Embedding on M\"{o}bius Ring}\label{sec:mobiusring}
	A KG is described by a set of triples $\Delta$, where each $l=(h,r,t)\in \Delta$ contains $h$, $r$, and $t$ as head
entity, relation, and tail entity, respectively. Denote $E=\{h, t | (h,r,t)\in\Delta\}$ and $R=\{r | (h,r,t)\in\Delta\}$
as the set of entities and relations of $\Delta$. Let $f$ be the scoring function, the task of KGE is to find vectors
$\mathbf{h}, \mathbf{r}, \mathbf{t}$ corresponding to $h,r,t$ which minimize:
\begin{eqnarray}
\mathcal{I} = \sum_{l=(h,r,t)\in \Delta}\sum_{(h',r,t')\in\Delta_l'}[\gamma + f(\mathbf{h},\mathbf{r},\mathbf{t}) -
f(\mathbf{h}',\mathbf{r},\mathbf{t}')]_{+}, \label{eq:KGEObj}
\end{eqnarray}
where
\begin{eqnarray}
H_l&=&\{(e,r,t) | e\in E, e\neq h\},\quad l=(h,r,t), \label{eq:HeadNegativeSample}\\
T_l&=&\{(h,r,e) | e\in E, e\neq t\},\quad l=(h,r,t), \label{eq:TailNegativeSample}\\
\Delta_{l}' &=& H_l\cup T_l - \Delta, \label{eq:TotalNegativeSample}
\end{eqnarray}
and $l=(h,r,t)\in \Delta$, $\gamma$ is a margin hyperparameter. The function $[\cdot]_{+}$ is defined by $[x]_{+}=\max(0, x)$ for $x\in\mathbb{R}$. 

Suppose $l=(h,r,t)$ be a positive triple, the trueness of $l$ assessed by $f$ is given by
\begin{eqnarray}
{\rm rank}_l &=& \mbox{The rank of } f(l) \mbox{ in the sequence of } \nonumber\\
& & \{f(l)\}\cup f(\Delta_{l}') \mbox{ in ascending order}. \label{eq:Rank}
\end{eqnarray}
In general, we use the value ${\rm rank}_l$ to evaluate the trueness of a candidate triplet. 

In order to begin with the following description, we give two critical functions $\mathbbm{m}_k(\cdot)$ and $\mathbbm{d}_k(\cdot)$. 

For any $u\in\mathbb{R}$ and $k\in\mathbb{Z}^+$, $\mathbbm{m}_k(u)$ is defined as\footnote{$\mathbbm{m}_1(u)$ is equivalent to $[u]$ in \cite{ebisu2018toruse}.}
\begin{eqnarray}
\mathbbm{m}_k(u)\in[0,k) \mbox{ and } u\equiv \mathbbm{m}_k(u)\,{\rm mod}\, k. 
\end{eqnarray}
$\mathbbm{d}_k(u)$ is defined as 
\begin{eqnarray}
\mathbbm{d}_k(u)=\min(\mathbbm{m}_k(u), k-\mathbbm{m}_k(u)).
\end{eqnarray}
Based on these definitions, for any $u\in \mathbb{R}$, it is not difficult to verify that $\mathbbm{m}_k(u)=\mathbbm{m}_k(u\pm k)$, $\mathbbm{m}_k(-u)=k-\mathbbm{m}_k(u)$, $\mathbbm{d}_k(-u)=\mathbbm{d}_k(u)$, and $\mathbbm{d}_k(u)=\mathbbm{d}_k(u\pm k)$. In particular, $\mathbbm{d}_k(u)$ can be reformulated as $\mathbbm{d}_k(u) = \min_{i\in \mathbb{Z}} |u+ik|$.

%-------------------------------------------------------------------------------------%
\subsection{Revisit Torus Ring $\mathbb{T}^{2}$}

\begin{figure}
  \centering
  \includegraphics[width=0.30\textwidth]{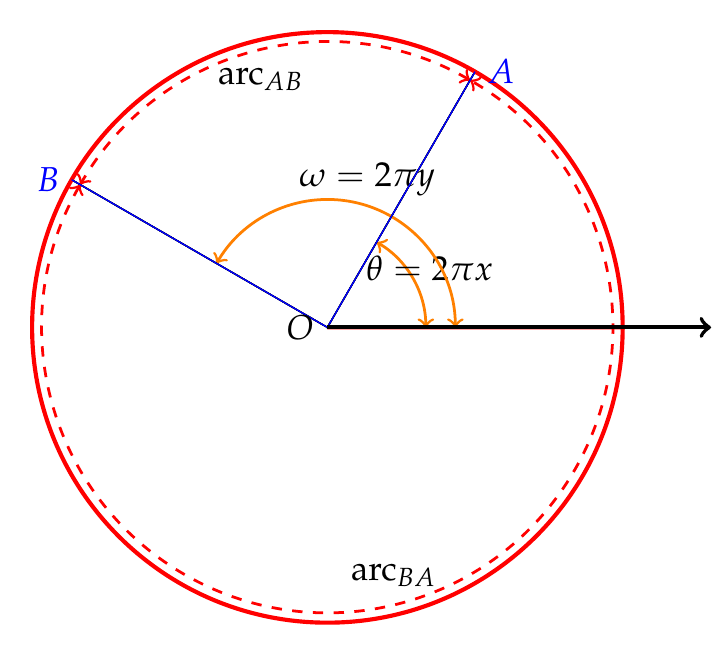}
  \caption{Illustration of Torus ring $\mathbb{T}^1$.} \label{fig:T1}
\end{figure}

Before introducing $\mathbb{T}^2$, we would like review the definition of Torus ring $\mathbb{T}^{1}$. As shown in Fig. \ref{fig:T1}, $\mathbb{T}^{1}$ is a one-dimensional ring. In particular, any two points $A$ and $B$ on $\mathbb{T}^{1}$ can be uniquely defined by their angles $\theta=2\pi x$ and $\omega=2\pi y$ with $x, y\in[0,1)$, these variables are all depicted in Fig. \ref{fig:T1}. 

The addition operation $\oplus$ of any two points $A$ and $B$ is defined by $x\oplus y = \mathbbm{m}_{1}(x+y)$. The distance between $A$ and $B$ can be viewed as the minimal value among the rotating angles from $A$ to $B$ and from $B$ to $A$, i.e., ${\rm arc}_{AB}$ and ${\rm arc}_{BA}$ as shown in Fig. \ref{fig:T1}. Describing such an idea in math, we have 
\begin{eqnarray}
{\rm dist}(x, y) = \min\{\mathbbm{m}_{1}(x-y), \mathbbm{m}_{1}(y-x)\} = \mathbbm{d}_{1}(x-y). \nonumber
\end{eqnarray}
This distance function (and all the following distance functions) satisfies
\begin{eqnarray}
\left\{\begin{array}{l}
\rm{dist}(x, x)=0, \\
\rm{dist}(x, y) = \rm{dist}(y, x)\geq 0, \\
\rm{dist}(x+y, z)\leq \rm{dist}(x, z) + \rm{dist}(y, z), 
\end{array}\right. \label{eq:distance_condition}
\end{eqnarray}
where $x, y, z\in [0,1)$. 

Torus ring $\mathbb{T}^2$ can be viewed as the independent stacking of two Torus ring $\mathbb{T}^1$. Given $x=(x_1,x_2), y=(y_1, y_2)\in [0,1)\times[0,1)$ as two points on Torus ring $\mathbb{T}^2$, then the addition operation on $\mathbb{T}^2$ is defined by 
$x\oplus y=(\mathbbm{m}_1(x_1+y_1), \mathbbm{m}_1(x_2+ y_2))$. The distance function on $\mathbb{T}^2$ is given as 
\begin{eqnarray}
\rm{dist}(x, y) = \|\digamma\|_{\tau}, \quad
\digamma = (\rm{dist}(x_1, y_1), \rm{dist}(x_2, y_2)), \nonumber
\end{eqnarray}
where $\|\cdot\|_\tau$ is some vector norm. 

Torus ring $\mathbb{T}^2$ can be extended to $n$-dimensional case, let $x=(x_i), y=(y_i)\in [0,1)^{\times n}$ as two points on $\mathbb{T}^n$, it holds
\begin{eqnarray}
x\oplus y &=&(\mathbbm{m}_1(x_1+y_1), \cdots, \mathbbm{m}_1(x_n+ y_n)), \label{eq:torusn-oplus} \\
\rm{dist}(x, y) &=& \|\digamma\|_{\tau}, \label{eq:torusn-distance},
\end{eqnarray}
where $\digamma=(\mathbbm{d}_1(x_1-y_1), \mathbbm{d}_1(x_2-y_2), \cdots, \mathbbm{d}_1(x_n-y_n)).$

Summarize the above discussion, the objective function (\ref{eq:KGEObj}) becomes TorusE when the scoring function $f(\cdot)$ is set as 
\begin{eqnarray}
f(\mathbf{h},\mathbf{r},\mathbf{t}) = {\rm dist}(\mathbf{h}\oplus\mathbf{r}, \mathbf{t}), \quad \mathbf{h},\mathbf{r},\mathbf{t}\in\mathbb{T}^n, \nonumber
\end{eqnarray} 
where ${\rm dist}(\cdot)$ is given by (\ref{eq:torusn-distance}) and $\oplus$ is given by (\ref{eq:torusn-oplus}).

%-------------------------------------------------------------------------------------%
\subsection{M\"{o}bius Ring $\mathbb{M}^{2}$}

\begin{figure}
	\centering
	\includegraphics[width=0.30\textwidth]{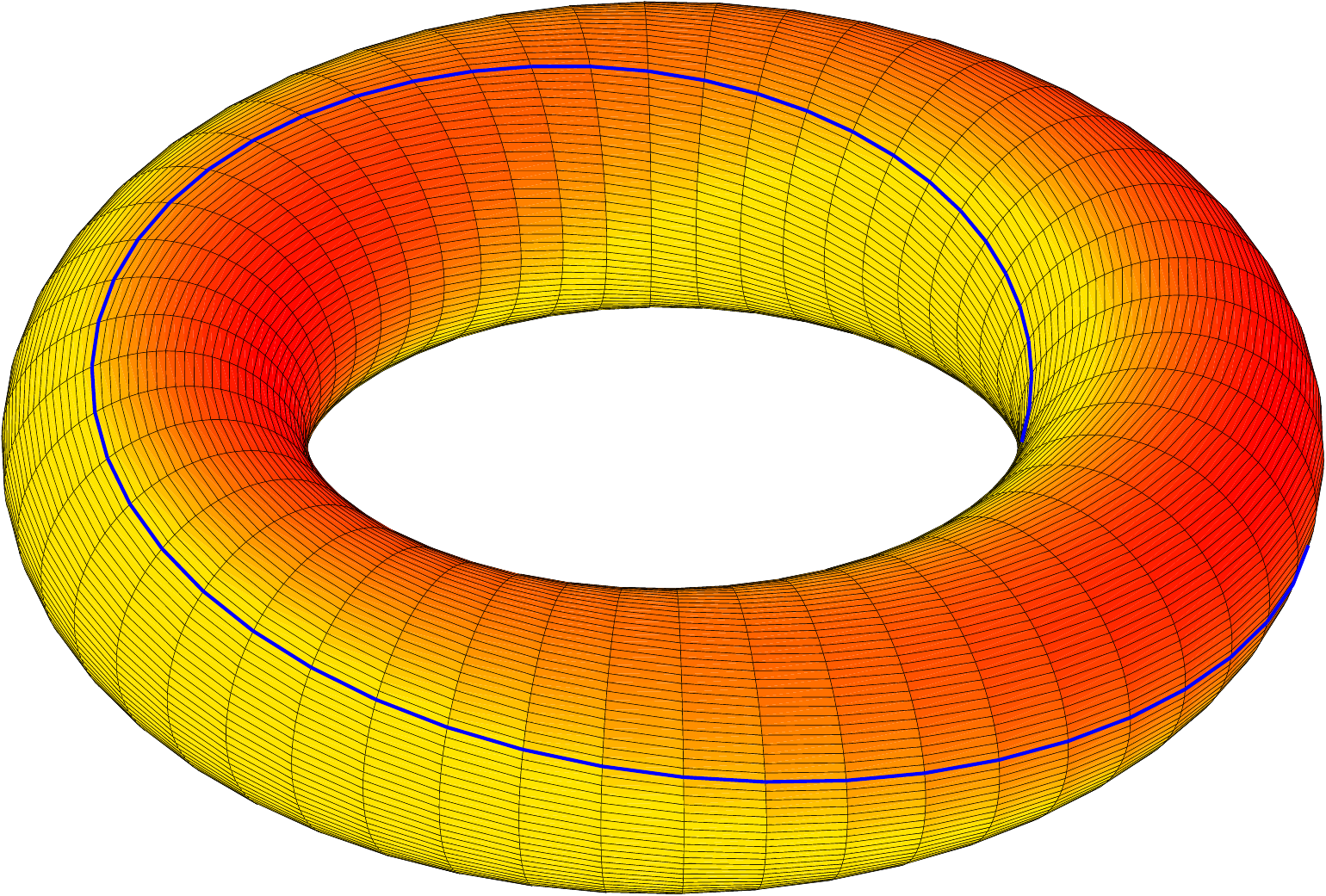}
	\caption{Parametirc curve on Mobius: $\omega=0$ in (\ref{eq:Mobius}).} \label{fig:mobius}
\end{figure}

As an intuitive illustration of M\"{o}bius ring, we plot the surface of M\"{o}bius ring $\mathbb{M}^{2}$ in Fig. \ref{fig:mobius}. In this figure, let the radius from the center of the hole to the center of the tube be $R$, the radius of the tube be $r$, then the parametric equation for M\"{o}bius ring is 
\begin{eqnarray}
\left\{\begin{array}{lcl}
x(\theta, \omega) &=& (R+r\cos(\frac{\theta}{2}+\omega))\cos\theta, \\
y(\theta, \omega) &=& (R+r\cos(\frac{\theta}{2}+\omega))\sin\theta, \\
z(\theta, \omega) &=& r\sin(\frac{\theta}{2}+\omega). \\
\end{array}\right. \label{eq:Mobius}
\end{eqnarray}
Letting $\theta = 2\pi x_1$ and $\omega = 2\pi x_2$, then any point on $\mathbb{M}^2$ can be uniquely defined by $(x_1, x_2)$. In particular, \emph{\color{red}the period of $x_1$ is $2$ and the period of $x_2$ is $1$}, hence we set $(x_1, x_2)\in [0,2)\times[0,1)$. 

Given $x=(x_1,x_2), y=(y_1, y_2)\in [0,2)\times[0,1)$ as two points on a M\"{o}bius ring, the addition operation $\oplus$ on $\mathbb{M}^2$ is defined by
\begin{eqnarray}
x\oplus y=(\mathbbm{m}_2(x_1+y_1), \mathbbm{m}_1(x_2+ y_2)), \label{eq:oplus-M2}
\end{eqnarray}
i.e., the addition on $\mathbb{M}^2$ is formulated by the modulus addition on each dimension with modulus $2$ and $1$, respectively. 

The distance function ${\rm dist}(\cdot)$ between $x$ and $y$ is defined as
\begin{eqnarray}\label{eq:distance-M2}
{\rm dist}(x, y) &=& \min(w_1, w_2), \\
w_1 &=& \mathbbm{d}_2(y_1-x_1)+\mathbbm{d}_1(y_2-x_2), \label{eq:distance-M2-w1} \\
w_2 &=& \mathbbm{d}_2(y_1-x_1+1)+\mathbbm{d}_1\Big(y_2-x_2+\frac{1}{2}\Big). \label{eq:distance-M2-w2}
\end{eqnarray}
One can further verify that the above definition follows the basic properties in (\ref{eq:distance_condition}). Moreover, the above distance satisfies: 

%-------------------------------------------------------------------------------------%
\begin{prop}\label{prop:distance_bound}
For distance defined in (\ref{eq:distance-M2}) for $\mathbb{M}^2$ and any two points $x=(x_1, x_2)$, $y=(y_1, y_2)$ in $[0,2)\times [0,1)$, it holds $0\leq {\rm dist}(x,y)\leq \frac{3}{4}$, where the upperbound $\frac{3}{4}$ achieves when $x_1-y_1=k\pm\frac{1}{2}$ and $x_2-y_2 =\frac{k'}{2}\pm\frac{1}{4}$ for some integer $k$ and $k'$. 
\end{prop}

We give the proof of {\em Proposition} \ref{prop:distance_bound} in Appendix \ref{appendix:equivalence}. In what follows, we will explain the reason for the definition (\ref{eq:distance-M2}) by using equation (\ref{eq:Mobius}). 

Given $x=(x_1, x_2), y=(y_1, y_2)\in \mathbb{M}^2$ and the corresponding points $A$ and $B$ on $\mathbb{M}^2$, transforming $A$ to $B$ (or, $B$ to $A$) needs one of the following two operations:

\begin{itemize}

\item[a).] If transform $A$ to $B$, then add $y_1-x_1 + 2k$ to the first component of $x$, and add $y_2-x_2+k'$ to the second component of $x$ for some integer $k$ and $k'$. If transform $B$ to $A$, then add $x_1-y_1 + 2k$ to the first component of $y$, and add $x_2-y_2+k'$ to the second component of $y$ for some integer $k$ and $k'$. Hence, when we measure the distance in the first dimension using modulus $2$ and considering that $x_1, y_1\in[0,2)$, such a distance is $\mathbbm{d_2}(y_1-x_1)$. Similarly, if we measure the distance between $x$ and $y$ in the second dimension using modulus $1$ and considering that $x_2, y_2\in[0,1)$, the obtained distance is $\mathbbm{d_1}(y_2-x_2)$. By adding the distance in the two dimensions, we simply obtain the distance in this case be $\mathbbm{d}_2(y_1-x_1)+\mathbbm{d}_1(y_2-x_2)$, which conforms with (\ref{eq:distance-M2-w1}).

\item[b).] If transform $A$ to $B$, then add $y_1-x_1 + 2k + 1$ to the first component of $x$, and add $y_2-x_2+k'+\frac{1}{2}$ to the second component of $x$ for some integer $k$ and $k'$. If transform $B$ to $A$, then add $x_1-y_1 + 2k + 1$ to the first component of $y$, and add $x_2-y_2+k'+\frac{1}{2}$ to the second component of $y$ for some integer $k$ and $k'$. In order to make an intuitive calculation, we calculate the value of $\frac{\theta}{2}+\omega$ and $\theta$ for $A$ in (\ref{eq:Mobius}) as  
$$\frac{\theta}{2}+\omega = \pi x_1 + 2\pi x_2, \quad \theta = 2\pi x_1,$$
when add $y_1-x_1 + 2k + 1$ to $x_1$, and add $y_2-x_2+k'+\frac{1}{2}$ to $x_2$, it becomes
\begin{eqnarray}
\theta &=& 2\pi y_1 + (4k+2)\pi, \nonumber\\
\frac{\theta}{2}+\omega &=& \pi (y_1+2k+1) + 2\pi (y_2+k'+\frac{1}{2}) \nonumber\\
&=&\pi y_1 + 2\pi y_2 + 2(k+k'+1)\pi, \nonumber
\end{eqnarray}
which coincides with the coordinate of $B$ on $\mathbb{M}^2$ (ignore multiple of $2\pi$ difference). Similar calculation can be conducted for the case of transforming $B$ to $A$ and hence we omit the details. To sum it up, when we measure the distance in the first dimension using modulus $2$ and considering that $x_1, y_1\in[0,2)$, such a distance is $\mathbbm{d_2}(y_1-x_1+1)$. Similarly, if we measure the distance between $x$ and $y$ in the second dimension using modulus $1$ and considering that $x_2, y_2\in[0,1)$, the obtained distance is $\mathbbm{d}_1(y_2-x_2+\frac{1}{2})$. By adding the distance in the two dimensions, we simply obtain the distance in this case be $\mathbbm{d}_2(y_1-x_1+1)+\mathbbm{d}_1(y_2-x_2+\frac{1}{2})$, which conforms with (\ref{eq:distance-M2-w2}). 
\end{itemize}

Summarize the above discussion, the distance between $x, y\in \mathbb{M}^2$ should be the minimum of the two cases, which is equivalent to (\ref{eq:distance-M2}). 

In fact, M\"{o}bius ring $\mathbb{M}^2$ degenerates to Torus ring in some special case, as shown in the following proposition, whose proof is given in Appendix \ref{appendix:equivalence}. 

%-------------------------------------------------------------------------------------%
\begin{prop}\label{prop:equivalence}
If the first dimension of $\mathbb{M}^{2}$ is set to zero, then $\mathbb{M}^{2}$ is equivalent to $\mathbb{T}^1$.
\end{prop}

For comparison between Torus ring and M\"{o}bius ring, we list the parametric equations for Torus ring $\mathbb{T}^2$ as: 
\begin{eqnarray}
\left\{\begin{array}{lcl}
x(\theta, \omega) &=& (R+r\cos\omega)\cos\theta, \\
y(\theta, \omega) &=& (R+r\cos\omega)\sin\theta, \\
z(\theta, \omega) &=& r\sin\omega. \\
\end{array}\right. \label{eq:Torus}
\end{eqnarray}

As we can see in (\ref{eq:Torus}), the period of $\theta$ and $\omega$ are both $2\pi$, hence the following points on $\mathbb{T}^2$ are equivalent:
\begin{eqnarray}
(\theta,\omega) \Leftrightarrow (\theta + 2k\pi, \omega+2k'\pi),\quad k, k'\in\mathbb{Z}. \nonumber
\end{eqnarray}
However, in M\"{o}bius (\ref{eq:Mobius}), the period of $\theta$ and $\omega$ are $4\pi$ and $2\pi$ respectively. In particular, the following points can be viewed as equivalent points on M\"{o}bius ring: 
\begin{eqnarray}
(\theta,\omega) &\Leftrightarrow& (\theta + 4k\pi, \omega+2k'\pi) \nonumber\\
&\Leftrightarrow& (\theta + (4k+2)\pi, \omega+(2k'+1)\pi),\quad k, k'\in\mathbb{Z}, \nonumber
\end{eqnarray}
which can be verified in (\ref{eq:Mobius}).

\begin{figure}
	\centering
	\includegraphics[width=0.30\textwidth]{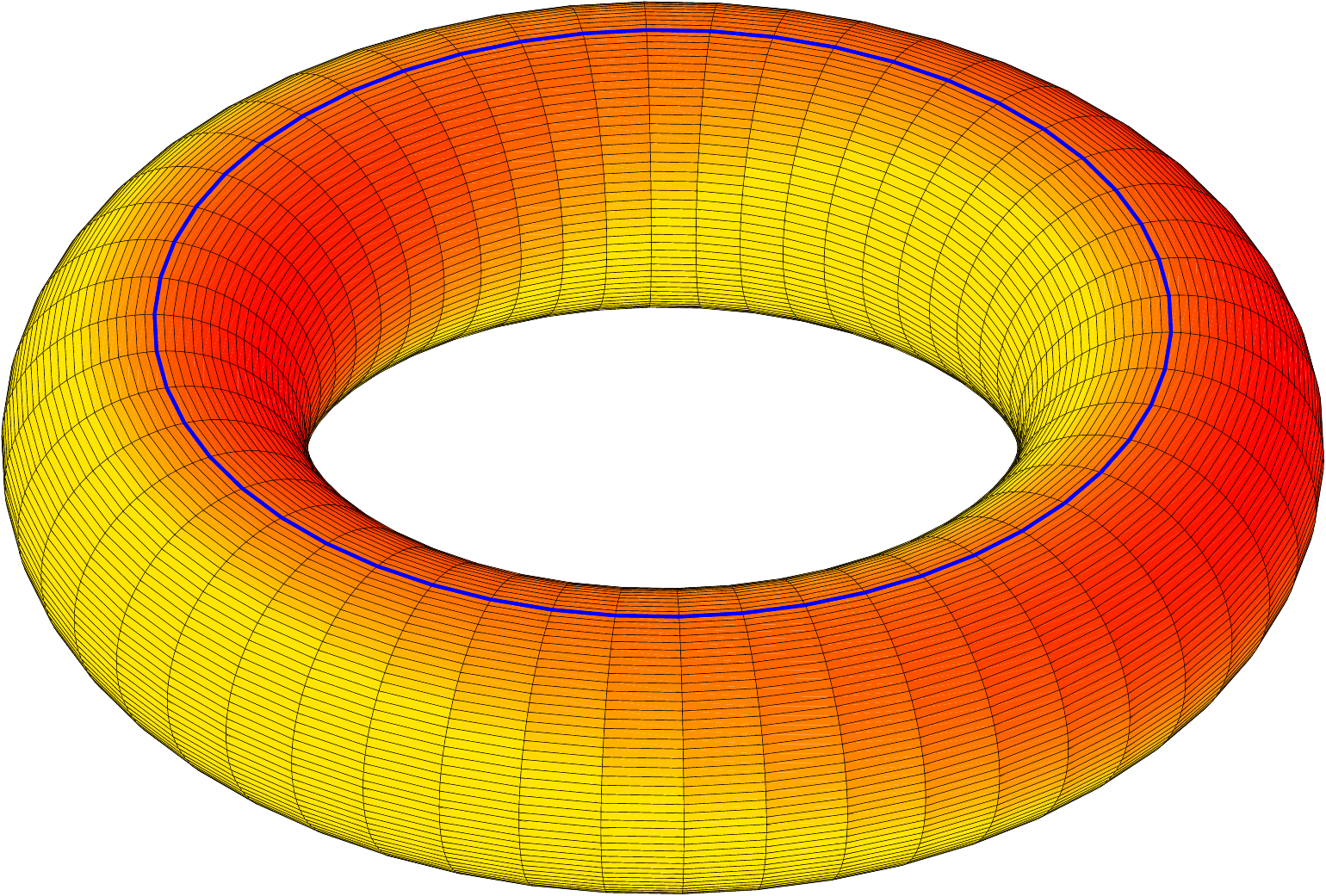}
	\caption{Parametirc curve on Torus: $\omega=\frac{\pi}{2}$ in (\ref{eq:Torus})} \label{fig:torus}
\end{figure}

As shown in Fig. \ref{fig:torus} and Fig. \ref{fig:mobius}, the parametric curve on Torus ring and M\"{o}bius ring are also quite different: fixing $\omega$ in (\ref{eq:Torus}) generates a cycle, but fixing $\omega$ in (\ref{eq:Mobius}) generates a twisted cycle. 

%-------------------------------------------------------------------------------------%
\subsection{M\"{o}bius Ring $\mathbb{M}^{q/p}$}

The simple M\"{o}bius ring $\mathbb{M}^{2}$ can be extended to more general case $\mathbb{M}^{q/p}$. M\"{o}bius ring $\mathbb{M}^{q/p}$ is a space constructed on $[0, q)\times [0,p)$, where $p, q$ are {\em\color{red}co-prime positive integers}, any $x,y\in \mathbb{M}^{q/p}$ satisfies:
\begin{eqnarray}
x\oplus y &=& (\mathbbm{m}_q(x_1+y_1), \mathbbm{m}_p(x_2+ y_2)), \\
\rm{dist}(x, y) &=& \min_{j=0}^{pq-1}\Big\{\mathbbm{d}_q(y_1-x_1+\frac{1}{p}j)+\nonumber\\
& & \mathbbm{d}_p(y_2-x_2+\frac{1}{q}j)\Big\}. \label{eq:distance-Mqp}
\end{eqnarray}
The reason of why $\rm{dist}(x, y)$ is defined as (\ref{eq:distance-Mqp}) is similar to that of (\ref{eq:distance-M2}) and hence omitted here. The distance $\rm{dist}(\cdot, \cdot)$ satisfies properties (\ref{eq:distance_condition}) too, and in particular there is:  

%-------------------------------------------------------------------------------------%
\begin{prop}\label{prop:Mqpdistance}
The distance defined in (\ref{eq:distance-Mqp}) for $\mathbb{M}^{q/p}$ satisfies 
$${\rm dist}(x,y)\in \Big[0,\frac{1}{2p}+\frac{1}{2q}\Big),$$
where the upperbound achieves when $(x,y)=\big(\frac{1}{2p}, \frac{1}{2q}\big)$.
\end{prop}

The proof of the above proposition is similar to that of {\em Proposition} \ref{prop:distance_bound} and hence omitted in this paper. The proof of $\rm{dist}(x+y, z)\leq \rm{dist}(x, z) + \rm{dist}(y, z)$ in $\mathbb{M}^{q/p}$ in (\ref{eq:distance_condition}) is given in Appendix \ref{sec:appendix2}.

In order to increase the embedding dimensions for KGE, we define
\begin{eqnarray}
\mathbb{M}^{q/p}_n = \underbrace{\mathbb{M}^{q/p} \uplus \mathbb{M}^{q/p} \uplus \cdots \uplus \mathbb{M}^{q/p}}_{n-1\;\;\uplus} \label{eq:Mqpuplus}
\end{eqnarray}
as the direct sum of $n$ M\"{o}bius ring $\mathbb{M}^{q/p}$. For each $u=(u_1, u_2, \cdots, $ $u_n)$, $v=(v_1, v_2, \cdots v_n)\in \mathbb{M}^{q/p}_n$, there is $v_i, u_i\in\mathbb{M}^{q/p}$ for each $1\leq i\leq n$ and 
\begin{eqnarray}
u\oplus v &=& (u_1\oplus v_1, \cdots, u_n\oplus v_n), \label{eq:oplus-Mqp}\\
{\rm dist}(u,v) &=& \|\digamma\|_{\tau}, \label{eq:distance-Mqp} 
\end{eqnarray}
where $\digamma = (\rm{dist}(u_1, v_1), \cdots, \rm{dist}(u_n, v_n))$ and $\|\cdot\|_{\tau}$ is some vector norm. 

{\em\color{red} For embedding on M\"{o}bius ring, the scoring function $f(\mathbf{h},\mathbf{r},\mathbf{t})$ in (\ref{eq:KGEObj}) will be set as
\begin{eqnarray}
f(\mathbf{h},\mathbf{r},\mathbf{t}) = {\rm dist}(\mathbf{h}\oplus\mathbf{r}, \mathbf{t}), \quad \mathbf{h},\mathbf{r},\mathbf{t}\in\mathbb{M}^{q/p}_{n}, \nonumber
\end{eqnarray}
where the operation $\oplus$ is defined in (\ref{eq:oplus-Mqp}) and distance function is defined in (\ref{eq:distance-Mqp})}, which is listed in Tab. \ref{models}.

In what follows, we list a key proposition for M\"{o}bius ring $\mathbb{M}^{q/p}$, whose proof is intuitive and hence omitted here. 

%-------------------------------------------------------------------------------------%
\begin{prop}\label{prop:zeropoints}
The solutions of the following equation 
\begin{eqnarray}
{\rm dist}(x, 0) = 0 \label{eq:zeropoint}
\end{eqnarray}
in the region $[0,q)\times[0,p)$ for $\mathbb{M}^{q/p}$ are $x=(iq/p, jp/q)$ with $0\leq i<p$ and $0\leq j<q$. 
\end{prop}

In the next, we call the solutions of equation (\ref{eq:zeropoint}) be zero points of $\mathbb{M}^{q/p}$.

For any given KG described by set of triplets $\Delta$, one of the {\em\color{red} difficulties for KGE comes from the cycle structure in the given KG data}, a cycle structure in KG means the existence of entities $h_i$ and relations $r_i$ ($0\leq i\leq m$) such that 
\begin{eqnarray}
(h_i, r_i, h_{i+1})\in \Delta \mbox{ or } (h_{i+1}, r_i, h_{i})\in \Delta \mbox{ for } 0\leq i\leq m-1,  \nonumber
\end{eqnarray}
and $(h_m, r_m, h_0) \in \Delta$ or $(h_0, r_m, h_m) \in \Delta$. In the case that the given set $\Delta$ contains no cycle structure, the geometric structure of $\Delta$ becomes a tree, and the embedding task for this KG without considering the negative samples can be implemented via a simple iteration strategy. However, realistic KG data contains a great amount of logic cycles and {\em\color{red} an appropriate KGE strategy should choose an algebraic structure which is capable of fitting the constraint generated by tremendous number of cycles. }

Consider the properties in (\ref{eq:distance_condition}), the constraint of the above cycle without entities can be described by an equation of relations on $\mathbb{M}^{q/p}$: 
\begin{eqnarray}
{\rm dist}(\gamma_0\hat{r}_0+  \gamma_1\hat{r}_1+ \cdots \gamma_s \hat{r}_{s}, 0) = 0, \label{eq:cycle}
\end{eqnarray}
where $\gamma_i\in \mathbb{Z}$, $\hat{r}_i\in \{r_0, r_1, \cdots, r_m\}$. 

In general, the number of cycles in a given KG dataset is much larger than that of the relations, hence the set of equations (\ref{eq:cycle}) in Euclidean space is hardly to have an exact solution. Intuitively, {\em\color{red}since the number of zero points in $\mathbb{M}^{q/p}$ in the region $[0,q)\times[0,p)$ is $pq$, the set of equations (\ref{eq:cycle}) derived by all the cycles in given KG data will have more chance to get an exact solution.} However, the number of zero point in $\mathbb{T}^2=[0,1)\times [0,1)$ is only $1$ for TorusE, which is much smaller than that of $\mathbb{M}^{q/p}$. In the above sense, M\"{o}biusE assumes more powerful expressiveness than TorusE. 

%---------------------------------------------------------------------------------------------------------%
\section{Experiments}\label{sec:experiments}

The performance of the proposed M\"{o}biusE is tested on two datasets: FB15K and WN18 \cite{bordes2013translating}, which are extracted from realistic knowledge graphs. The basic properties on these two datasets is given in Tab. \ref{tab:data}.

We conduct the link prediction task by using the same method reported in \cite{bordes2013translating}. For each test triplet (which is a positive sample), we replace its head (or tail) to generate a corrupted triplet, then the score of each corrupted triplet is calculated by the scoring function $f(\cdot)$, and the ranking of the test triplet is obtained according to these scores, i.e., the definition of ${\rm rank}_l$ in (\ref{eq:Rank}). It should be noted from (\ref{eq:TotalNegativeSample}) that the set of generated corrupted elements excludes the positive triplets $\Delta$, we call such a set be ``filtered'' \cite{ebisu2018toruse}, and the ranking values in our experiments are all obtained on the ``filtered'' set. 

We choose stochastic gradient descent algorithm to optimize the objective (\ref{eq:KGEObj}). In order to generate the negative sample in (\ref{eq:KGEObj}), we employ the ``Bern'' method \cite{wang2014knowledge} for negative sampling because of the datasets only contains positive triplets. 

To evaluate our model, we use Mean Rank (MR), Mean Reciprocal Rank (MRR) and HIT@\emph{m} as evaluating indicators \cite{bordes2013translating}. MR is calculated by ${\rm mean}_{l\in \Delta}({\rm rank}_{l})$ with $\Delta$ be the set of triplet data and ${\rm rank}_l$ is defined in (\ref{eq:Rank}), MRR is calculated by ${\rm mean}_{l\in \Delta}(1/{\rm rank}_{l})$, HIT@\emph{m} is defined by $|\{l\in\Delta: {\rm rank}_l\leq m\}|/|\Delta|$. 

We conduct a grid search to find a set of optimal hyper-parameters for each dataset, the searching area for margin $\gamma$ in (\ref{eq:KGEObj}) is $\{2000,1000,500,200,100\}$ and the searching area for learning rate $\alpha$ in stochastic gradient descent is $\{0.002, 0.001, $ $0.0005, 0.0002, 0.0001\}$. Finally, the learning rate $\alpha$ is set to $0.0005$ both for WN18 and FB15K, the margin $\gamma$ is set to $2,000$ for WN18 and $500$ for FB15K. 

We conduct our experiments on the augmented M\"{o}bius ring $\mathbb{M}^{q/p}_n$ as defined in (\ref{eq:Mqpuplus}), and we choose three different types of M\"{o}bisu ring for embedding, i.e., $\mathbb{M}^{2/1}_n$, $\mathbb{M}^{3/1}_n$, $\mathbb{M}^{3/2}_n$. In choosing the distance function in $\mathbb{M}^{q/p}_n$, we choose $\|\cdot\|_{\tau}=\|\cdot\|_1$ in ${\rm dist}(\cdot)$ in (\ref{eq:distance-Mqp}). We call the above corresponding M\"{o}biusE be M\"{o}biusE(2,1), M\"{o}biusE(3,1), M\"{o}biusE(3,2) in Tab. \ref{tab:15K} and Tab. \ref{tab:wn18}. The value of $n$ in $\mathbb{M}^{q/p}_n$ is set to $5,000$, which means in each embedding vector we have $10,000$ parameters to be trained. The dimensions for other types of models in Tab. \ref{tab:15K} and Tab. \ref{tab:wn18} are all set to $10,000$. 

As shown in Tab. \ref{tab:15K}, M\"{o}bius(3,1) outperforms the other models in 4 of the 5 critical indicators. In Tab. \ref{tab:wn18}, M\"{o}bius(2,1) outperforms the other models also in 4 of the 5 critical indicators. 

\begin{table}
	\small
	\centering
	\begin{tabular}{c|ccccc}
		\toprule
		Dataset 	& \#Ent 	&\#Rel	&\#Train	&\#Valid	&\#Test \\
		\midrule
		WN18  	&40,943		&18		&141,442	&5,000	&5,000\\
		FB15K  	&14,951		&1,345	&483,142	&50,000	&59,071\\
		\bottomrule
	\end{tabular}
		\caption{Statistics of tested datasets.}
	\label{tab:data}
\end{table}

\begin{table}
	\small
	\centering
	\begin{tabular}{lccccc}
		\toprule
		Model&MRR&MR&HIT@10&HIT@3&HIT@1 \\
		\midrule
		TransE  &0.516&\bf{73.21}&0.847&0.738&0.269\\
		TransH  &0.559&82.48&0.795&0.680&0.404\\
		RESCAL & 0.354 & - & 0.587 & 0.409 & 0.235\\
		DistMult &0.690&151.40&0.818&0.749&0.609\\
		ComplEx &0.673&210.17&0.796&0.713&0.606\\
		TorusE &0.746&143.93&0.839&0.784&0.689\\
		\midrule
		M\"{o}biusE(2,1)&0.782& 118.53&0.862&\bf{0.817}&0.731\\
		M\"{o}biusE(3,1)&\bf{0.783}& 114.50&\bf{0.863}&\bf{0.817}&\bf{0.734}\\
		M\"{o}biusE(3,2)& 0.767& 73.87&\bf{0.863}& 0.816& 0.702\\
		\bottomrule
	\end{tabular}
		\caption{Performance of M\"{o}biusE in FB15K}
	\label{tab:15K}
\end{table}

\begin{table}
	\small
	\centering
	\begin{tabular}{lccccc}
		\toprule
		Model&MRR&MR&HIT@10&HIT@3&HIT@1 \\
		
		\midrule
		TransE  &0.414&-&0.688&0.534&0.247\\
		TransR  &0.218&-&0.582&0.404&0.218\\
		RESCAL & 0.890 & - & 0.928 & 0.904 & 0.842\\
		DistMult &0.797&655&0.946&-&-\\
		ComplEx &0.941&-&0.947&0.945&0.936\\
		TorusE &\bf{0.947}&577.62&\bf{0.954}&\bf{0.950}&0.943\\
		\midrule
		M\"{o}biusE(2,1)&\bf{0.947}& 539.80&\bf{0.954}&\bf{0.950}&\bf{0.944}\\
		M\"{o}biusE(3,1)&\bf{0.947}& 531.80&\bf{0.954}&\bf{0.950}& 0.943\\
		M\"{o}biusE(3,2)&0.944&\bf{425.51}&\bf{0.954}& 0.949& 0.937\\
		\bottomrule
	\end{tabular}
\caption{Performance of M\"{o}biusE in WN18}
\label{tab:wn18}
\end{table}

%---------------------------------------------------------------------------------------------------------%
\section{Related Works}\label{sec:relatedworks}

The key for different types of KGE strategies is the choosing of embedding spaces, and in turn different addition operations and different scoring functions, i.e., different $f(\cdot)$ in (\ref{eq:KGEObj}). As the first work in KGE, TranE \cite{bordes2013translating} uses the basic algebraic addition to generate the score function, and regularizes the obtained vectors via a normalized condition. In TransE, the left (right) entity and the relation uniquely defines the right (left) entity. However, for realistic KG, it may have the following properties: a). ONE-TO-MANY mapping, given any fixed relation $r$, the left (right) entity corresponding to a the right (left) entity is not unique; b): MANY-TO-ONE mapping, given any fixed entities $h$ and $t$, the relation between $h$ and $t$ may not be unique. The original TransE cannot resolve the above two drawbacks in the beginning.

As an improvement of TransE to overcome the above drawback a), TransR \cite{lin2015learning} maps each relation $r$ to a vector and a matrix, the corresponding scoring function is derived by the image of such a relation matrix. In this sense, different entities may be mapped to a same vector in the image space and the drawback a) of TransE is resolved. Similar to TransR, TransH \cite{wang2014knowledge} uses a single vector to obtain the image of each entities. As further generalization of TransR and TransH, TransD \cite{ji2015dynamicmapping} uses a dynamic matrix which is determined both by relation and entities to generate the above mapping matrix. As another solution to the drawbacks of TransE, TransM \cite{fan2014transitionbased} introduces a relation-based adjust factor to the scoring function in TransE, such a factor will be much smaller in MANY-TO-MANY case than that of ONE-TO-ONE case and hence penalizing effects works. 

Introducing flexibility in the scoring function in some extent enhances the generalization capability of KGE. As an implementation of such an idea, TransF \cite{feng2015flexible} uses a flexible scoring function in KGE, in which the matching degree is desribed by the inner products of desired entities and given relation. Increasing the complexity in the scoring function enhances the nonlinear representative capability of KGE, RESCAL \cite{nickel2011rescal} and DistMult \cite{yang2015embedding} uses a matrix-induced inner product to represent the scoring function in KGE, and ComplEx \cite{trouillon2012complex} applies the similar but embeds both relation and entities to a complex-valued space. TransA \cite{jia2018knowledge} combines the idea of TransE and RESCAL and incorporates the residue vector in TransE to the vector-based norm in RESCAL, such a strategy can adaptively find the loss function according to the structure of knowledge graphs. 

M\"{o}biusE is inspired from TorusE \cite{ebisu2018toruse}, however, M\"{o}biusE differs from TorusE in the following aspects: The minimal dimension of Torus ring is $1$, but the minimal dimension of M\"{o}bius ring is $2$; The addition in Torus ring chooses an unique value for modulus operation (generally $1$), but the addition in M\"{o}bius ring chooses different values (generally $p$ and $q$ in $\mathbbm{M}^{q/p}$); The distance function is M\"{o}bius ring is strongly nonlinear (see, (\ref{eq:distance-Mqp})), which is much more complicated than that of Torus ring. These major difference guarantees the strong expressiveness of M\"{o}biusE. 

%---------------------------------------------------------------------------------------------------------%
\section{Conclusions}\label{sec:conclusions}

A novel KGE strategy is proposed by taking advantage of the intertwined rotating property of M\"{o}bius ring. As a first step, we defined the basic addition operation on M\"{o}bius ring and discussed its related properties. Next, in order to obtain an appropriate scoring function based on M\"{o}bius ring, we built a distance function on M\"{o}bius ring and constructed the corresponding distance-induced scoring function. Finally, a complete KGE strategy is obtained via the above constructions, which outperforms several key KGE strategies in our conducted experiments. 

%---------------------------------------------------------------------------------------------------------%
\section{Appendix}\label{sec:appendix}

%-------------------------------------------------------------------------------------%
\subsection{Proof of $\rm{dist}(x, z)\leq \rm{dist}(x, y) + \rm{dist}(y, z)$ in $\mathbb{M}^{q/p}$. }\label{sec:appendix2}

According to the definition of $\rm{dist}(\cdot, \cdot)$ on $\mathbb{M}^{q/p}$, we only need to prove
$\rm{dist}(x+y, 0)\leq \rm{dist}(x, 0) + \rm{dist}(y, 0)$.

Note that for any $i_1, i_2, i_1', i_2', j, j'\in \mathbb{Z}$, there is
\begin{eqnarray}
& &\mathbbm{d}_q(y_1+x_1+\frac{1}{p}j)+\mathbbm{d}_p(y_2+x_2+\frac{1}{q}j) \nonumber\\
&\leq &|y_1+x_1+\frac{1}{p}j+i_1q|+|y_2+x_2+\frac{1}{q}j+i_2p| + \nonumber\\
&\leq & |x_1+\frac{1}{p}j+\frac{1}{p}j'+i_1q + i_1'q| + \nonumber\\
&     & |x_2+\frac{1}{q}j+\frac{1}{q}j'+i_2p+i_2'p| + \nonumber\\
&     & |y_1-\frac{1}{p}j'-i_1'q|+|y_2-\frac{1}{q}j'-i_2'p| \nonumber
\end{eqnarray}
based on which and the arbitrariness of $i_1$ and $i_2$, there is
\begin{eqnarray}
&     & {\rm dist}(x+y,0) \nonumber\\
&    =& \min_{j=0}^{pq-1}\left\{\mathbbm{d}_q(y_1+x_1+\frac{1}{p}j)+\mathbbm{d}_p(y_2+x_2+\frac{1}{q}j)\right\}\nonumber\\
&\leq & \min_{j=0}^{pq-1}\left(\min_{i_1}|x_1+\frac{1}{p}j+\frac{1}{p}j'+i_1q + i_1'q|\right. + \nonumber\\
&     & \left.\min_{i_2}|x_2+\frac{1}{q}j+\frac{1}{q}j'+i_2p+i_2'p|\right) + \nonumber\\
&     & |y_1-\frac{1}{p}j'-i_1'q|+|y_2-\frac{1}{q}j'-i_2'p| \nonumber\\
&=    & \min_{j=0}^{pq-1} \left(\mathbbm{d}_q(x_1+\frac{1}{p}(j+j'))+\mathbbm{d}_p(x_2+\frac{1}{q}(j+j'))\right)\nonumber\\
&     & + |y_1-\frac{1}{p}j'-i_1'q|+|y_2-\frac{1}{q}j'-i_2'p|. \nonumber\\
& =   & {\rm dist}(x,0) + |y_1-\frac{1}{p}j'-i_1'q|+|y_2-\frac{1}{q}j'-i_2'p|. \nonumber
\end{eqnarray}

Due to the arbitrariness of $j', i_1', i_2'$, there is ${\rm dist}(x+y,0) \leq {\rm dist}(x,0) + {\rm dist}(y,0)$. 

%-------------------------------------------------------------------------------------%
\subsection{Proof of Proposition \ref{prop:distance_bound}}\label{appendix:distance_bound}
We define a function $g(\alpha, \beta)=\min\{g_1, g_2\}$ with $g_1 = \mathbbm{d}_2(\alpha) + \mathbbm{d}_1(\beta)$, $g_2 = \mathbbm{d}_2(\alpha+1) + \mathbbm{d}_1(\beta+\frac{1}{2})$. Based on which we know $\alpha$ has a period of $2$ and $\beta$ has a period of $1$. Next, we divide the interval $[-1,0)$ to $4$ subintervals $I_{\alpha, i}=[-1+0.5i, -0.5+0.5i)$, and divide $[-0.5,0.5)$ to $4$ subintervals $I_{\beta, j}=[-0.5+0.25j, -0.25+0.25j)$ with $i,j=0,1,2,3$, then the value of $\sup_{x,y\in I_{\alpha_i}\times I_{\beta,j}}g$ can be calculated and all these $16$ values are $3/4$.

\subsection{Proof of Proposition \ref{prop:equivalence}}\label{appendix:equivalence}
Let $x=(0, x_2)$, $y=(0, y_2)$, then ${\rm dist}(x,y)=\min(\mathbbm{d}_1(y_2-x_2), 1+\mathbbm{d}_1(y_2-x_2+\frac{1}{2}))=\mathbbm{d}_1(y_2-x_2)$, hence $\mathbb{M}^{2}$ can be viewed as $\mathbb{T}^1$ by constraining the first dimension to zero.

%\begin{itemize}
%\item document style
%\item baselineskip
%\item front matter
%\item keywords and MSC codes
%\item theorems, definitions and proofs
%\item lables of enumerations
%\item citation style and labeling.
%\end{itemize}%

%\begin{enumerate}[(1)]
%\item Group the authors per affiliation.
%\item Use footnotes to indicate the affiliations.
%\end{enumerate}

%\section*{References}

\bibliography{MobiusE}

\end{document}